\documentclass[10pt,twocolumn,letterpaper]{article}

\usepackage[pagenumbers]{cvpr} 

\usepackage{graphicx}
\usepackage{amsmath}
\usepackage{amssymb}
\usepackage{booktabs}

\usepackage{inconsolata}
\usepackage{pifont}
\usepackage{bm}
\usepackage{balance}

\usepackage[pagebackref=true,breaklinks=true,colorlinks,bookmarks=false]{hyperref}

\hypersetup{
	plainpages=false,
	colorlinks=true,              %
	linkcolor=blue,               %
	anchorcolor=blue,             %
	citecolor=blue,               %
	filecolor=blue,               %
	urlcolor=blue,                %
	pdfview=FitH,                 %
	pdfstartview=FitH,            %
	pdfpagelayout=SinglePage      %
}

\usepackage[capitalize]{cleveref}
\crefname{section}{Sec.}{Secs.}
\Crefname{section}{Section}{Sections}
\Crefname{table}{Table}{Tables}
\crefname{table}{Tab.}{Tabs.}

\def\cvprPaperID{*****} 
\def\confName{CVPR}
\def\confYear{2022}

\begin{document}

\title{\textsc{UniCon+}: ICTCAS-UCAS Submission to the\\AVA-ActiveSpeaker Task at ActivityNet Challenge 2022}
\author{Yuanhang Zhang\thanks{Equal contribution.} $^{1,2}$, Susan Liang$^{*1,2}$, Shuang Yang$^{1,2}$, Shiguang Shan$^{1,2}$\\
$^1$Key Laboratory of Intelligent Information Processing of Chinese Academy of Sciences (CAS),\\ Institute of Computing Technology, CAS, Beijing 100190, China\\
$^2$School of Computer Science and Technology, University of Chinese Academy of Sciences,\\ Beijing 101408, China\\
{\tt\small\{zhangyuanhang15, liangsusan18\}@mails.ucas.ac.cn, \{shuang.yang, sgshan\}@ict.ac.cn}
}
\maketitle

\begin{abstract}
   This report presents a brief description of our winning solution to the AVA Active Speaker Detection (ASD) task at ActivityNet Challenge 2022. Our underlying model \textsc{UniCon+} continues to build on our previous work, the Unified Context Network (UniCon)~\cite{DBLP:conf/mm/ZhangLYLWSC21} and Extended UniCon~\cite{zhangictcas} which are designed for robust \rm{scene-level} \it ASD. We augment the architecture with a simple GRU-based module that allows information of recurring identities to flow \rm{across scenes} \it through read and update operations. We report a best result of $94.47$\% mAP on the AVA-ActiveSpeaker test set, which continues to rank first on this year's challenge leaderboard and significantly pushes the state-of-the-art.\footnote{Project website: \url{https://unicon-asd.github.io/}.}
\end{abstract}

\section{Proposed Approach}
Our approach builds on an earlier work~\cite{zhangictcas}. In the next section, we summarize the newly proposed \textsc{UniCon+} architecture and training setups used for the challenge.
\label{sec:proposed}
\subsection{\textsc{UniCon+}}
\begin{figure*}
    \centering
    \includegraphics[width=\textwidth]{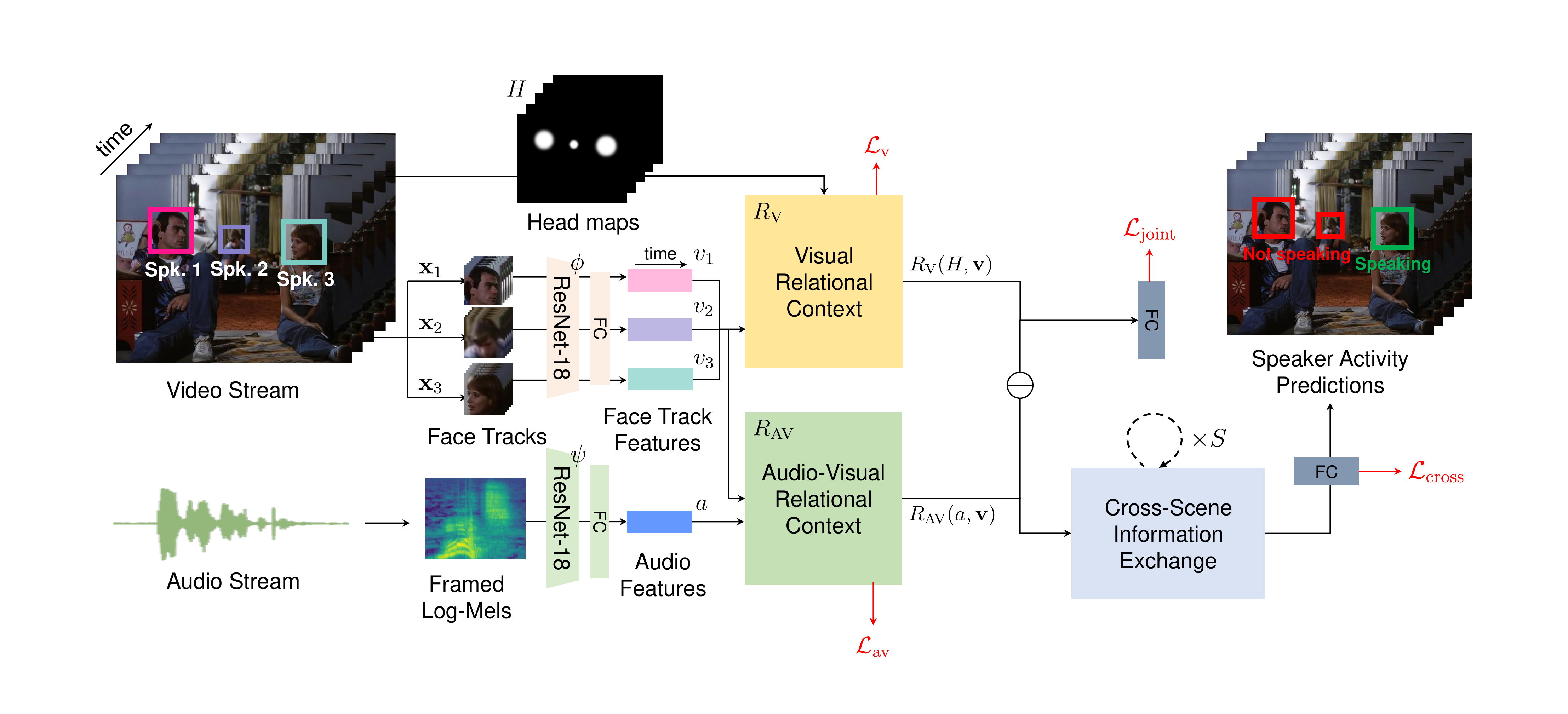}
    \vskip-2ex
    \caption{\textbf{Model overview.} We augment our previous work, Extended UniCon with a GRU-based cross-scene information exchange module, which learns to leverage information beyond the current scene for recurring identities.}
    \label{fig:overview}
\end{figure*}

In this subsection, we outline the design of our \textsc{UniCon+} model. Fig.~\ref{fig:overview} provides a graphic overview.
\paragraph{Scene (shot)-level modeling:} The main components are inherited from Extended UniCon~\cite{zhangictcas}, which combines multiple types of contextual information to optimize all candidates within a scene (shot) jointly. We employ two 2D ResNet-18s~\cite{DBLP:conf/cvpr/HeZRS16} for visual and audio feature extraction, a VGG-like convolutional encoder for head map encoding, and a $3$-layer Conformer~\cite{DBLP:conf/interspeech/GulatiQCPZYHWZW20} for temporal context modeling. For details, please refer to~\cite{DBLP:conf/mm/ZhangLYLWSC21,zhangictcas}. The only difference is the kernel size of the convolutions in the Conformer modules, which is set to $7$.
\paragraph{Information aggregation:} After obtaining each candidate's contextual visual and audio-visual features $R_{\text{V}}$ and $R_{\text{AV}}$ (see \cite{zhangictcas} for clarification on the notations), as well as scene-level active speaker scores $\bm p=(p_1,p_2, \dots, p_N)$ where $N$ is the number of candidates in the scene, we aggregate \textit{identity-} and \textit{speech-related} and information for each candidate as follows:
\begin{align}
    R_i &= R_{\text{V},i}\oplus R_{\text{AV},i},\\\label{eq:joint}
    \mathrm{Id}_i &= \frac 1T\sum_{t=1}^TR_{i,t},\\
    \mathrm{Sp}_i &= \frac{\sum_{t=1}^T p_{i,t}R_{i,t}}{\varepsilon+\sum_{t=1}^T p_{i,t}},
\end{align}
where $\oplus$ denotes concatenation, $T$ is the number of time steps, $i=1,2,\dots, N$ and $\varepsilon$ is a small value that prevents numerical overflow. Here, identity-related information is a plain unweighted average over all time steps, while speech-related information is weighed using the initial scene-level active speaker probabilities.
\paragraph{Cross-scene information exchange:} We use bidirectional Gated Recurrent Units (Bi-GRUs) to implement a cross-scene information exchange module. First, each identity's history information is initialized with a common learnable vector, which also serves as the initial input to the GRU. When the model sees a new scene, each candidate's aggregated speech- and identity-related information are concatenated and fed to the GRU, and the corresponding identity's history information is updated with the outputs of the GRU for the current scene.

Formally, let $K$ be the number of unique identities in the video, $S$ be the total number of scenes, $\bm h_0=(h_{1,0}=h_{\text{init}},\dots,h_{K,0}=h_{\text{init}})=\bm h_{S+1}$ be the pool of initial history vectors, and $\varphi (s,i)$ be the mapping from face track ID to person ID. The process can then be described as:
\begin{align}
    s_{i,s}&=\mathrm{Id}_{i,s}\oplus\mathrm{Sp}_{i,s},\\
    \bm s_s&=(s_{1,s},\dots,s_{N_s,s}),\\
    \bm h_s&=\big(h_{\varphi(s,1),s},\dots,h_{\varphi(s,N_s),s}\big),\\
    \overrightarrow{\bm s_s^{'}}&=\textsf{GRU}\big(\bm h_{s-1}, \overrightarrow{\bm s_{s-1}^{'}}\big),\label{eq:forward}\\
    \overleftarrow{\bm s_s^{'}}&=\textsf{GRU}\big(\bm h_{s+1}, \overleftarrow{\bm s_{s+1}^{'}}\big),\\
    \bm s_s^{'}&=\overrightarrow{\bm s_s^{'}}+\overleftarrow{\bm s_s^{'}}.
\end{align}
where $N_s$ is the number of candidates in scene $s$, $s=1,2,\dots, S$ and $i=1,2,\dots, N_s$. In our experiments, the Bi-GRU has $2$ layers and $576$ cells per direction per layer.

Finally, each scene's exchanged information $\bm s_s'$ is concatenated on a per-frame basis to the raw joint features $R$ (Eq.~\eqref{eq:joint}), and passed through two fully-connected layers to produce the final active speaker probabilities.
\paragraph{Loss function:} To reflect the addition of the above information exchange module, we augment the original loss function with a term $\mathcal L_{\textrm{cross}}$ to supervise the newly added classifier. Again, we apply the standard binary cross-entropy (BCE) loss, averaged over all time steps. The total loss hence becomes:
\begin{equation}
    \mathcal{L}=\mathcal{L}_{\textrm{av}}+\mathcal{L}_{\textrm{v}}+\mathcal{L}_{\textrm{joint}}+\mathcal{L}_{\textrm{cross}}.
\end{equation}
\subsection{Ensembling and Test-Time Augmentation}
\label{subsec:ensemble_tta}
To further boost performance, we perform ensembling in terms of model averaging, and apply test-time augmentation. For model averaging, we average the weights from the five best-performing checkpoints on the validation set to produce the final checkpoint. For test-time augmentation, we crop each face track into four corner tubes and the central tube plus the flipped version of these ($10$ crops in total). To obtain the final active speaker probabilities, we run inference on each of these augmented versions, average the ten resulting raw logits, and finally pass the result through sigmoid activation. Indeed, this proves to be very useful when potentially important face regions lie outside central crops due to extreme poses or bounding box drift.
\subsection{Implementation Details}
\label{subsec:impl}
We carry out our experiments on the large-scale AVA-ActiveSpeaker dataset~\cite{DBLP:conf/icassp/RothCKMGKRSSXP20} which consists of $262$ YouTube movies from film industries around the world. To obtain identity information for the face tracks, we leverage the recently released Audiovisual Person Search (APES) dataset~\cite{DBLP:conf/cvpr/AlcazarCMPLAG21}, which assigns a unique identity to each face track in the AVA-ActiveSpeaker training and validation set. In particular, we treat face tracks annotated as \texttt{EXTRA\_OR\_AMBIGUOUS} as unique, unrelated identities. For the test set, we do not have ground truth identity mappings, so we obtain approximate identities via face track clustering. Specifically, for each face track, face embeddings are extracted offline for every frame and then averaged over time. The network used is an off-the-shelf ResNet-100~\cite{DBLP:conf/cvpr/HeZRS16} ArcFace model~\cite{DBLP:conf/cvpr/DengGXZ19,An_2022_CVPR} pre-trained on the Glint360k dataset\footnote{\url{https://github.com/deepinsight/insightface/tree/master/recognition/partial_fc}}. The resulting embeddings are clustered via Agglomerative Hierarchical Clustering (AHC)~\cite{day1984efficient}, with cannot-link constraints applied to overlapping face tracks by adding large distance penalties. Note that we prefer slight under-clustering to over-clustering since cluster purity is crucial.

Our data preprocessing scheme is identical to that described in \cite{DBLP:conf/mm/ZhangLYLWSC21}. We implement our model with PyTorch~\cite{DBLP:conf/nips/PaszkeGMLBCKLGA19} and the \texttt{pytorch-lightning} package. All models are trained from scratch, using the AdamW optimizer~\cite{loshchilov2018decoupled} and automatic mixed precision (AMP) on NVIDIA V100 GPUs, each with 32GB memory. The network parameters are initialized using He initialization~\cite{DBLP:conf/iccv/HeZRS15}. We use a $60$-epoch cosine learning rate schedule, warming up linearly to a maximum learning rate of $0.0005$ (when training on one candidate or multiple scenes) or $0.0001$ (when training on multiple candidates) over the first $10$ epochs and decaying thereafter. When training on multiple scenes, every example within a batch consists of up to four sampled scenes which contain a common identity. Early stopping is applied when suitable.

During training, we augment the visual data via random horizontal flipping and uniform corner cropping along the input face tracks, followed by random adjustments to brightness, contrast, and saturation. We augment the audio tracks with SpecAugment~\cite{DBLP:conf/interspeech/ParkCZCZCL19}. All cropped face tracks are resized to $144\times 144$, and randomly cropped to $128\times 128$ for training.
\section{Results}
\label{sec:results}
The official metric for the task is Mean Average Precision (mAP). We obtain the numbers using the official evaluation tool, after interpolating our predictions to the timestamps in the original annotations. The results are shown in Table \ref{table:ava-sota}.

\begin{table*}[ht]
	\renewcommand{\arraystretch}{1.2}
	\centering
		\begin{tabular}{l|c|c|c}
			\hline
			\textbf{Method (Organization)} & \textbf{Val mAP (\%)} & \textbf{Test mAP} (\%) & \textbf{Pre-training?} \\
			\hline\hline
			AV-GRU (Google baseline) \cite{DBLP:conf/icassp/RothCKMGKRSSXP20}
			& $82.2$ & $82.1$ & \ding{55}\\ 
			Multi-Task (UCAS) \cite{zhangmulti2019} 
			& $84.0$& $83.5$ & \ding{51}\\
			ASC (Universidad de los Andes) \cite{DBLP:conf/cvpr/AlcazarCMPLAG20,alcazaruniversidad}
			& $87.1$ & $86.7$ & \ding{51} \\ 
			VGG-\{LSTM,TempConv\} (Naver Corporation) \cite{DBLP:journals/corr/abs-1906-10555}
			& $87.8$ & $87.8$ & \ding{51}\\
			MAAS-TAN \cite{DBLP:conf/iccv/AlcazarHTG21}
			& $88.8$ & N/A & \ding{51} \\ 
			SPELL\cite{DBLP:journals/corr/abs-2112-01479} & $90.6$ & N/A & \ding{51} \\
			EESEE-2D\cite{DBLP:journals/corr/abs-2203-14250} & $91.1$ & N/A & \ding{51} \\
			Uncertainty Fusion\cite{DBLP:conf/interspeech/PouthierPGBP21} & $91.9$ & $89.5$ & \ding{55} \\
			UniCon\cite{DBLP:conf/mm/ZhangLYLWSC21,zhangictcas} & $92.0$ & $90.7$ & \ding{55} \\
			TalkNet (National University of Singapore)\cite{DBLP:conf/mm/TaoPDQS021,taonus} & $92.3$ & $90.8$ & \ding{55} \\
			ASD-Transformer\cite{Datta2022} & $93.0$ & N/A & \ding{55} \\
			ASDNet (Technical University of Munich)\cite{DBLP:conf/iccv/KopukluTR21,kopukluasdnet} & $93.5$ & $91.9$ & \ding{51} \\
			Extended UniCon \cite{zhangictcas} & $93.6$ & $93.3$ & \ding{55} \\
			Extended UniCon$^\dagger$ (ICTCAS-UCAS-TAL)\cite{zhangictcas} & $93.8$ & $93.4$ & \ding{55} \\
			EESEE-50 (IVUL-KAUST)\cite{DBLP:journals/corr/abs-2203-14250} & $94.1$ & $93.0$ & \ding{51} \\
			\textbf{Ours (\textsc{UniCon+}), per-scene inference (w/o exchange)} & $94.5$ & $94.1$ & \ding{55}\\\hline
			\textbf{Ours (\textsc{UniCon+}), best$^\dagger$}
			& $\bf 94.7$ & $\bf 94.5$ & \ding{55} \\ \hline
		\end{tabular}
	\caption{Comparison with previous work on AVA-ActiveSpeaker. $^\dagger$: ensembling and/or test-time augmentation.}
	\label{table:ava-sota}
\end{table*}

Surprisingly, from the table we see that \textsc{UniCon+} can outperform our previous winning solution \cite{zhangictcas} even when each scene is evaluated individually, i.e. no cross-scene information exchange is performed. We explain this result as the trained GRU module being able to help extract more pertinent information, even for a single scene, in which case the scene-level representations are enhanced. Applying multi-scene inference and the practices described in Sec.~\ref{subsec:ensemble_tta}, we achieve a new state-of-the-art of $94.5$\% mAP on the test set.

In addition, we continue to outperform recent competitive models that employ heavier 3D or hybrid 3D-2D visual backbones, and without pre-training. Our model also remains end-to-end trainable with a reasonably large batch size. Inference can be performed quite efficiently in a single pass: on a single NVIDIA V100 GPU, it only takes about $10$ minutes for the entire end-to-end pipeline to finish processing the $5.8$-hour validation set (batch size is set to $1$).

\section{Discussions}
\begin{figure}
    \centering
    \begin{subfigure}{0.48\linewidth}
    \includegraphics[width=0.95\linewidth]{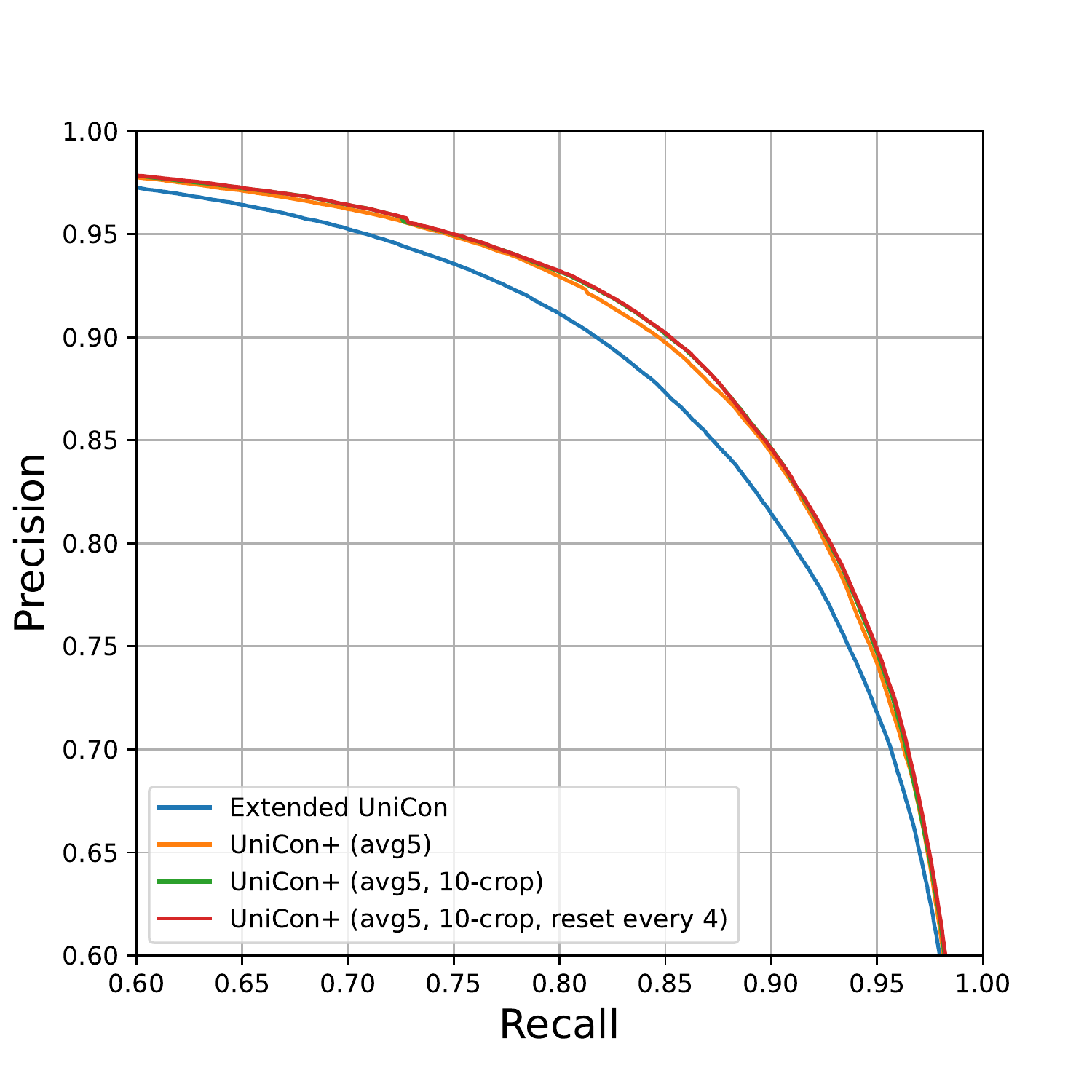}
    \caption{Precision-Recall}
    \end{subfigure}
	\hfill
	\begin{subfigure}{0.48\linewidth}
    \includegraphics[width=0.95\linewidth]{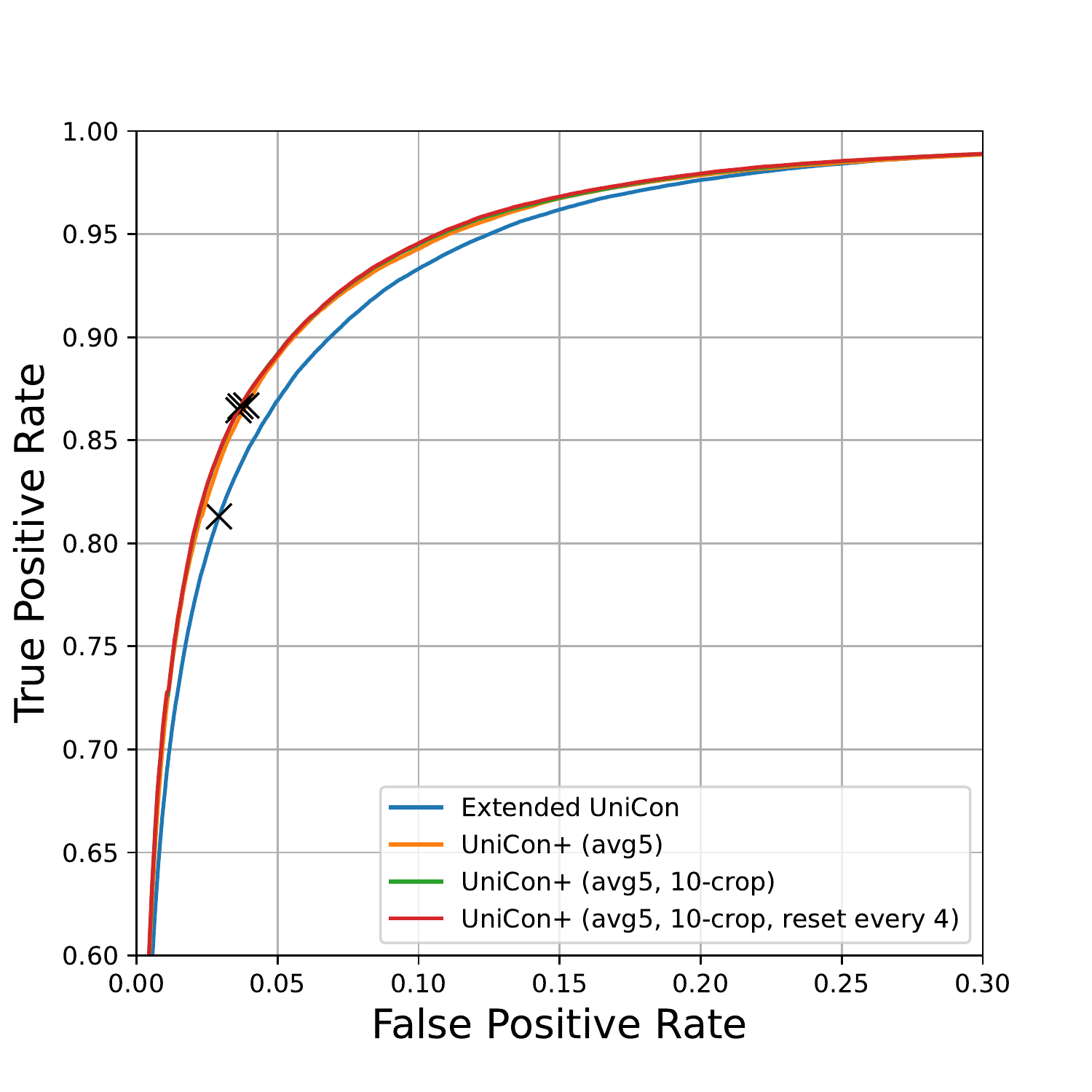}
    \caption{ROC}
    \end{subfigure}
	\caption{Precision-Recall and ROC curves for selected models (see legends).}
	\label{fig:pr_roc}
\end{figure}

\begin{figure}
    \centering
    \includegraphics[width=0.75\linewidth]{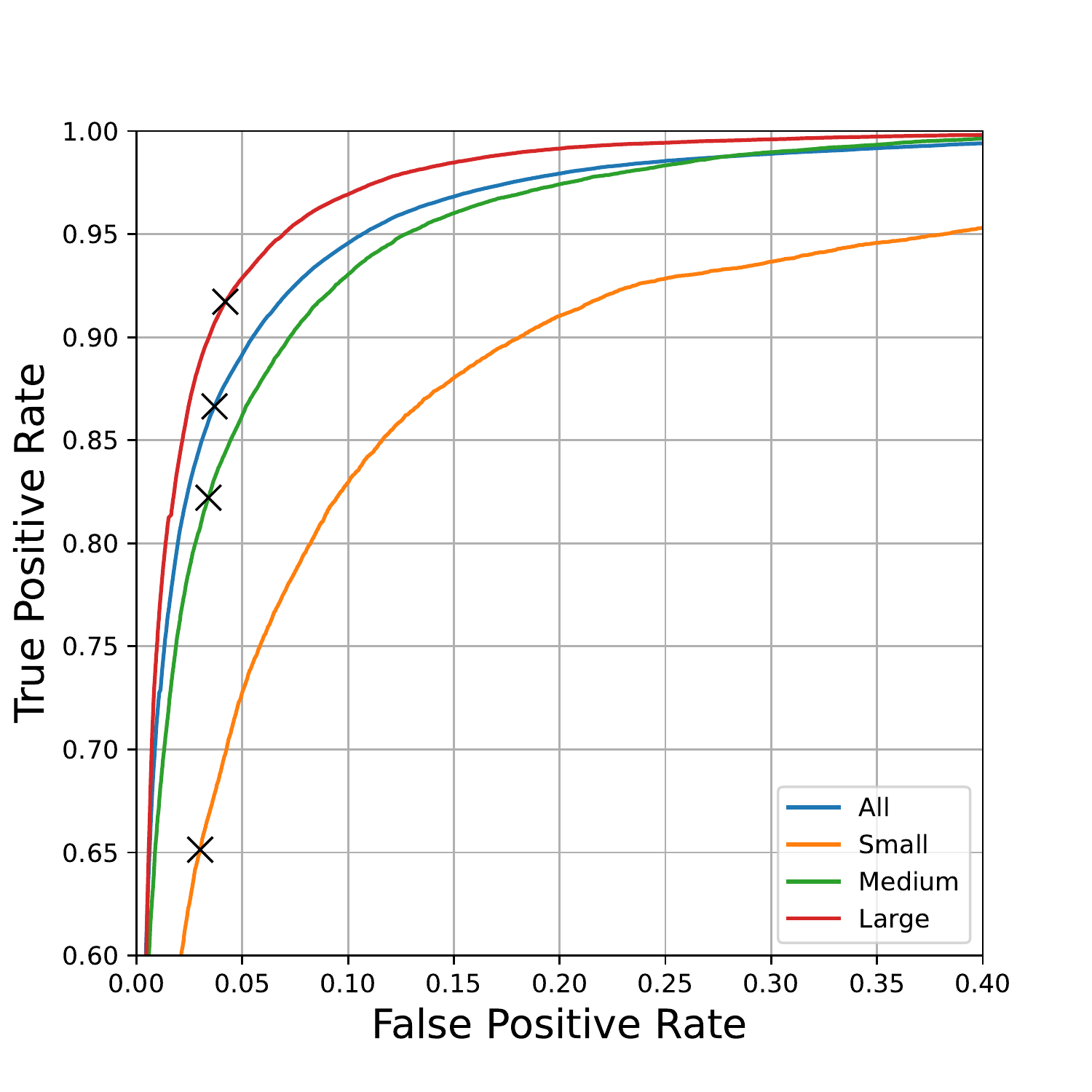}
	\caption{ROC curves for the best model (corresponding to the last row in Table \ref{table:ava-sota}), partitioned by face sizes.}
	\label{fig:breakdown}
\end{figure}

We plot the Precision-Recall and ROC curves on the validation set for a few selected models in Fig.~\ref{fig:pr_roc}. The $\times$ in the ROC curves represents the $p=0.5$ balanced accuracy point. Fig.~\ref{fig:breakdown} additionally shows ROC curves for the best submission, partitioned by face size: small ($<64$ pixels wide), medium ($>64$, $<128$ px) and large ($>128$ px), similar to the one in \cite{DBLP:conf/icassp/RothCKMGKRSSXP20}. We note that performance on the ``large" partition is nearly saturated, and for different face sizes, FPR at the balanced accuracy point is almost constant, showing that the resulting model still works for scenes of different resolutions without careful calibration and tuning. However, there is still a notable gap in terms of TPR for the ``small" partition.

Finally, as mentioned in Sec.~\ref{subsec:impl}, when using face clusters instead of oracle track identity information, we prefer under-clustering to promote cluster purity. In practice, we find that doing this alone was not enough to combat noises within the clusters, and may even hurt performance. Moreover, although we use Bi-GRUs for information exchange, for this submission we only leveraged past information during inference (Eq.~\eqref{eq:forward}), which can be problematic if early updates are unreliable. Therefore, we additionally reset the memory state for each identity every four occurrences. The current design choice clearly leaves much room for improvement, which is left as future work.

\section*{Acknowledgements}
This work was partially supported by the National Key R\&D Program of China (No. 2017YFA0700804) and the National Natural Science Foundation of China (No. 61876171, 62076250).
{\small
\bibliographystyle{ieee_fullname}
\balance
\bibliography{egbib}
}

\end{document}